\documentclass[conference]{IEEEtran}
\usepackage{times,amsmath,epsfig,cite}
\title{Efficient Design of Triplet Based Spike-Timing Dependent Plasticity}

\author{%
{Mostafa Rahimi Azghadi, Said Al-Sarawi, Nicolangelo Iannella, and Derek Abbott}
\vspace{1.6mm}\\
\fontsize{10}{10}\selectfont\itshape
Centre for Biomedical Engineering, School of Electrical and Electronic Engineering,\\
The University of Adelaide, Adelaide, SA 5005, Australia\\\fontsize{9}{9}\selectfont\ttfamily\upshape
\{mostafa,alsarawi,iannella,dabbott\}@eleceng.adelaide.edu.au
\fontsize{10}{10}\selectfont\rmfamily\itshape
}
\begin{document}
\maketitle

\begin{abstract}
Spike-Timing Dependent Plasticity (STDP) is believed to play an important role in learning and the formation of computational function in the brain. The classical model of STDP which considers the timing between pairs of pre-synaptic and post-synaptic spikes (p-STDP) is incapable of reproducing synaptic weight changes similar to those seen in biological experiments which investigate the effect of either higher order spike trains (e.g. triplet and quadruplet of spikes)~\cite{ref2,ref3,ref5}, or, simultaneous effect of the rate and timing of spike pairs~\cite{ref4} on synaptic plasticity. In this paper, we firstly investigate synaptic weight changes using a p-STDP circuit~\cite{ref6} and show how it fails to reproduce the mentioned complex biological experiments. We then present a new STDP VLSI circuit which acts based on the timing among triplets of spikes (t-STDP) that is able to reproduce all the mentioned experimental results. We believe that our new STDP VLSI circuit improves upon previous circuits, whose learning capacity exceeds current designs due to its capability of mimicking the outcomes of biological experiments more closely; thus plays a significant role in future VLSI implementation of neuromorphic systems.         
\end{abstract}


\section{Introduction}
Numerous studies have shown that timing and patterns of spikes play an important role in changing the strength of synapses, a phenomenon called Spike Timing-Dependent Plasticity (STDP). Presented results in~\cite{ref1} suggest that this rule plays an important role in learning and memory in the brain. Previous experiments have demonstrated that the classical form of STDP which just considers the timing difference between pairs of pre-synaptic and postsynaptic spikes (p-STDP) fails to reproduce several biological experiments including those that consider higher order spike trains (e.g. triplets and quadruplets of spikes)~\cite{ref2,ref3,ref5}, and those that in addition to timing difference between pairs of spikes, bring the rate of spike-pairs into action of changing the synaptic weight~\cite{ref4}. In 2006, Pfister and Gerstner reported a new STDP rule, based upon the triplets of spikes (t-STDP), which better replicates the above mentioned experimental outcomes~\cite{ref2}. Since timing-based synaptic changes can play an important role underlying adaptive changes to the internal connectivity structure of spiking neural networks, many studies have been conducting physical implementation of these rules in recent years~\cite{ref6,ref7,ref8,ref9,rahimitriplet,ramakrishnan2011floating}.

This paper proposes a new VLSI implementation for t-STDP. This implementation is based on an improved version of a p-STDP circuit already presented in~\cite{ref6} which is able to reproduce the exponential behaviour of the learning window, but it fails to generate some important biological experiments presented in~\cite{ref3,ref4,ref5}. We show that the proposed t-STDP circuit succeeds in reproducing all the experimentally observed biological effects with reduced normalized mean square error. 

The remainder of this paper is organized as follows. In section~\ref{sec:pair}, p-STDP model and its corresponding circuit are presented and described. Section~\ref{sec:triplet} introduces the new t-STDP circuit and describes how it is related to the t-STDP learning rule. Section~\ref{sec:exp} provides some information about experimental setup including, experimental protocols, data sets and error function. Simulation results are provided in section ~\ref{sec:sim}. Section~\ref{sec:prev} compares the proposed design to previous works, followed by discussion and conclusion in sections~\ref{sec:discussion} and~\ref{sec:conc}.

\section{Pair-based STDP rule and circuit}\label{sec:pair}
\begin{figure*}
\centering
  \includegraphics[width=1\textwidth]{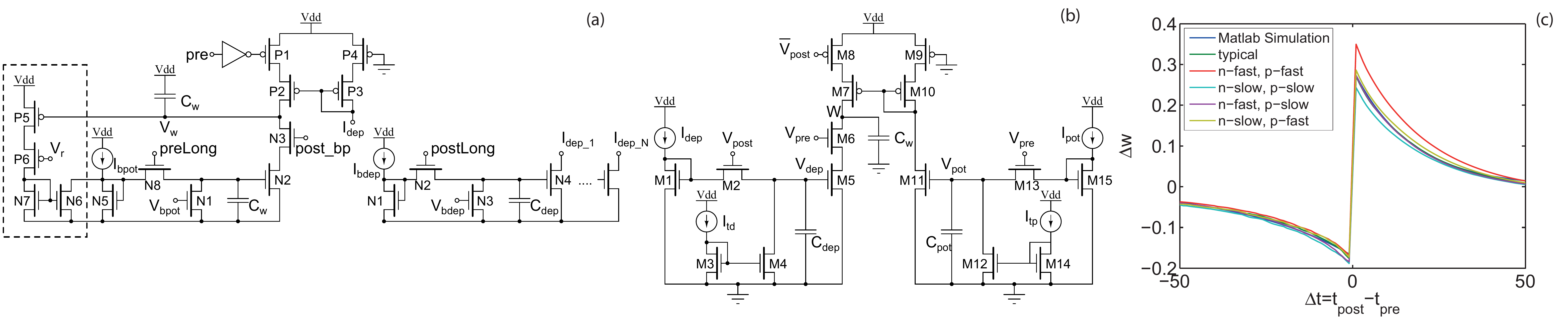}
  \caption{(a) p-STDP circuit presented in~\cite{ref6}. (b) Modified p-STDP circuit.(c) Pair-based STDP learning window generated by Matlab simulation and also using different transistor process corners for the modified p-STDP circuit shown in part (b) (Similar protocols and time constants to~\cite{ref1} were employed).}\label{fig:1}
\end{figure*}

The p-STDP synaptic modification rule governs weight changes in synapse based on the timing difference between pairs of spikes Fig.~\ref{fig:1}(a). This weight change dynamic can be given as:

\begin{equation}\label{eq:stdp}
\Delta w = \begin{cases} \Delta w^+=A^+e^{\rm (\frac{-\Delta t}{\tau_+})} & \mbox{if}~\Delta t\geq0 \\ \Delta w^-=-A^-e^{\rm (\frac{\Delta t}{\tau_-})} & \mbox{if}~\Delta t<0~, \end{cases}
\end{equation}
where $\Delta t=t_{\rm post}-t_{\rm pre}$ is the time difference between a single pair of post- and pre-synaptic spikes, $\tau_+$ and $\tau_-$ are time constants of the learning window, and $A^+$ and $A^-$ represent the maximal weight changes for potentiation and depression, respectively.  

There are several VLSI implementations that try to mimic the dynamics of the p-STDP rule~\cite{ref6,ref7,ref8}. The circuits presented in~\cite{ref7,ref8} cannot reproduce the required exponential behaviour seen in Eq.~\ref{eq:stdp}, while the circuit presented in~\cite{ref6} (Fig.~\ref{fig:1}(a)) can. 
Since the targeted t-STDP rule in this paper has an exponential behaviour, the circuit presented in~\cite{ref6} would be a more suitable candidate for implementing the t-STDP model. However, in order to make the circuit in~\cite{ref6} compatible with the t-STDP rule, some modifications are needed, leading to the circuit shown in Fig.~\ref{fig:1}(b). The modifications are as follows: (i) Since the classical p-STDP model (Eq.~\ref{eq:stdp}) is weight independent, in the modified circuit, the weight dependence part (shown in the dashed box in Fig.~\ref{fig:1}(a)) is omitted. (ii) Potentiation and depression in the modified circuit are represented with increased, and reduced amount of charges stored on the weight capacitor, respectively, which is in contrast to the circuit presented in~\cite{ref6}. (iii) Also, in order to simplify the circuit, preLong and postLong pulses which should be generated by an additional circuitry, were replaced with~$V{\rm pre}$ and~$V{\rm post}$. These signals represent the input pre- and post-synaptic pulses in the modified circuit. (iv) Furthermore, when considering the implementation aspects of the circuit presented in~\cite{ref9}, using bias voltages for times constants control results in significant variation in time constants under various 3$\sigma$ process corners. So, in order to make the circuit more robust against this condition, the bias voltages were represented as the gate-to-source voltage of a number of diode connected transistors that are biased by current sources (M3 and M14).
Fig.~\ref{fig:1}(c) demonstrate that using this approach results in very slight changes in time constants for all different process corners when compared with MATLAB simulations. 
Furthermore, these current sources ($I_{\rm pot}$ and $I_{\rm dep}$) can later be used to fine tune the time constants when needed. 
This approach suggests that the time constants, as well as the amplitude parameters are more robust against process mismatch.
In order to facilitate the simulation of these circuit, the scaling approach used in similar VLSI implementations of synaptic plasticity in~\cite{ref9,ref10} was used, which is microseconds instead of milliseconds, i.e a scale factor of 1000. However, in all simulation results presented in this paper, the results are scaled back to biological time to make the comparison easier. 
       
\section{Triplet-based STDP rule and circuit}\label{sec:triplet}

Triplet-based STDP (t-STDP), as its name infers, changes the synaptic weight based on the timing difference among triplets of spikes. According to the triplet rule presented in~\cite{ref2}, half of the potentiation and depression interactions in the triplet rule are similar to what happens in the p-STDP model. It means that if a pre-synaptic spike arrives in a specified time window after a post-synaptic spike, the synaptic weight should decrease, and it should increase if the reverse order of spike pairs happens. However, the second half is where the non-linearity of the t-STDP model appears. According to the triplet rule, when a pre-synaptic spike happens, it not only interacts with its previous post-synaptic spike(s), but also with its succeeding pre-synaptic spike(s) as well. The same scenario goes for the time a post-synaptic spike happens and it has effect on the previous pre-synaptic as well as post-synaptic spike(s). 

It is worth mentioning that, the t-STDP model is not just a simple change in the degree of freedom of the p-STDP model, but it tries to overcome some deficiencies of the p-STDP model. In~\cite{ref2}, it is addressed that the t-STDP model, removed two main problems of the p-STDP formula. These problems and how the t-STDP solves them are as follows: (1) As p-STDP considers just pairs of spikes, for any value of~$A^+>0$, if a pre-synaptic spike precedes a post-synaptic one, it brings about potentiation, while according to~\cite{ref4}, at low repetition frequencies, there is no potentiation. In the t-STDP model, this deficiency can be solved by setting~${A_2}^+$ to a small value or in the case of the minimal rule to zero. So it makes the potentiation very small which can be neutralized by a bit of depression, or it can be zero~\cite{ref2}. (2) Considering biological experiments in~\cite{ref4}, for ~$\Delta_t>0$, potentiation will increase with the increase in frequency. However, this behaviour cannot be generated by p-STDP, as when the frequency of pairs of spikes increases, it causes the pairs to interact with each other, so it causes no significant potentiation. This problem can be solved again by correct tuning the t-STDP parameters. In this case,~${A_3}^+$ should be strong enough to make the potentiation wins over depression and so to have depression in high frequencies~\cite{ref2}.  
       
\begin{figure}
\centering
  \includegraphics[width=0.5\textwidth]{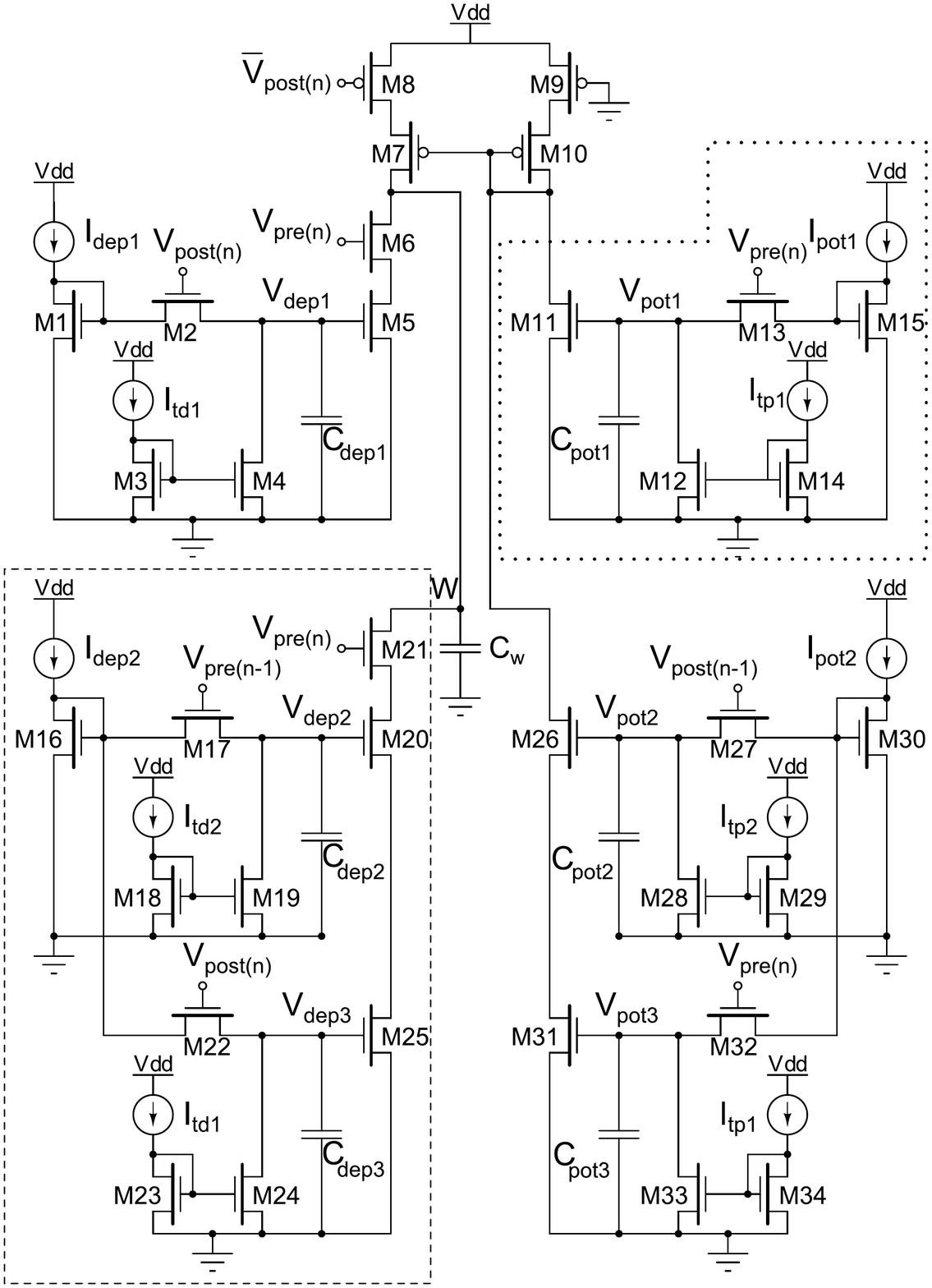}
  \caption{Proposed full Triplet-based STDP circuit.}\label{fig:triplet}
\end{figure}

Fig.~\ref{fig:triplet} presents the proposed circuit implementation of the t-STDP model. In the proposed circuit, there are eight parameters that can be tuned by controlling eight bias currents as follows: $I_{\rm dep1}$, $I_{\rm pot1}$, $I_{\rm dep2}$ and $I_{\rm pot2}$ represent the amplitude of synaptic weight changes for post-pre, pre-post, pre-post-pre and post-pre-post combinations of spike triplets, respectively. Another control parameter for these amplitude values in the circuit is the pulse width of the spikes which was kept fixed during all experiments in this paper (1 $\mu s$). In addition to these amplitude parameters, the required time constants in the model for post-pre, pre-post, pre-post-pre and post-pre-post combinations of spike triplets, can be adjusted using $I_{\rm td1}$, $I_{\rm tp1}$, $I_{\rm td2}$ and $I_{\rm tp2}$ respectively. 
 
The proposed circuit works as follows: upon the arrival of a post-synaptic pulse, $V_{\rm post(n)}$, M2, M8 and M22 switched on. At this time, $I_{\rm dep1}$ can charge the first depression capacitor, $C_{\rm dep1}$, through M2 to the voltage of $V_{\rm dep1}$. After finishing $V_{\rm post(n)}$, $V_{\rm dep1}$ starts decaying linearly through M4 and with a rate proportional to $I_{\rm td1}$. Now, if a pre-synaptic pulse, $V_{\rm pre(n)}$ arrives at M6 in the decaying period of $V_{\rm dep1}$, namely when M5 is still active, the weight capacitor, $C_W$, will be discharged through M5-M6 transistors and a depression occurs during the presence of a pre-synaptic pulse in the interval of affect of a post-synaptic spike (post-pre combination of spikes). Additionally, if a pre-synaptic spike arrives at M13, soon before the present post-synaptic spike at M8, the weight capacitor can be charged through M7-M8 transistors and a potentiation happens. This potentiation happens because the current post-synaptic spike is in the time of affect of a pre-synaptic spike (pre-post combination of spikes). The amount of potentiation depends on $V_{\rm pot1}$, which itself can be tuned by the relevant amplitude parameter $I_{\rm pot1}$. Also, the activation interval of M11 can be modified by changing the related time constant parameter $I_{\rm tp1}$. Furthermore, another contribution to potentiation can occur if a previous post-synaptic pulse, $V_{\rm post(n-1)}$, arrives at M27 soon enough before the current post-synaptic happens at M8 and also before a pre-synaptic pulse happens at M32 (this is the same pulse as for M13). In this situation, the weight capacitor can be charged again through M7-M8 and by an amount proportional to $V_{\rm pot2}$ and $V_{\rm pot3}$. This is a triplet interaction in the proposed circuit that leads to the required nonlinearity mentioned in the triplet learning rule, appears. A similar description holds for the situation when a pre-synaptic pulse occurs at M6, M13 and M21 transistors. But this time one potentiation and two depression events can happen if the appropriate situation is provided. 

The first two parts of the t-STDP circuit (on the top left and the top right) are identical to the p-STDP circuit presented in Fig.~\ref{fig:1}(b). Also, the two bottom parts of this circuit carry out the triplet terms interactions. This circuit is in correspondence to the full triplet learning rule presented in~\cite{ref2} which takes into account all four possible potentiation and depression interactions. However, as it is shown in following sections, only some of the terms are really necessary to reproduce the expected biological experiments. This is referred to as minimal triplet learning rule in~\cite{ref2} which makes the required circuit much simpler and smaller.              

\section{Experiments setup}\label{sec:exp}

In order to validate the functionality of the pair- and triplet-based circuits in comparison with experimental data observed in biological experiments, experimental protocols, data sets, and error function identical to those presented in~\cite{ref2}, are adopted. Four required experimental protocols, two different sets of data and the error function are explained in the following.
 
\subsection{Experimental protocols}\label{subsec:exp}
\subsubsection{Pairing protocol}\label{subsec:exp-pair}
Pair-based STDP protocol has been extensively used in electrophysiological experiments and simulation studies~\cite{ref2}. In this protocol, 60 pairs of pre- and post-synaptic spikes with a delay of $\Delta t$ are repeated with repetition frequency of $\rho$~Hz. In many experiments $\rho=1$~Hz, however, it has been illustrated in~\cite{ref4} that how altering the repetition frequency affects the total change in weight of the synapse.

\subsubsection{Triplet protocol}\label{subsec:exp-triplet}
There are two types of triplet patterns which are used in this paper. Both of them consist of 60 triplets of spikes which are repeated at a given frequency of $\rho=1$~Hz. The first triplet pattern is composed of two presynaptic spikes and one post-synaptic spike in a pre-post-pre configuration. As a result, there are two delays between the first pre and the middle post, $\Delta t_1=t_{post}-t_{pre1}$, and between the second pre and the middle post $\Delta t_2=t_{post}-t_{pre2}$. The second triplet pattern is analogous to the first but with two postsynaptic spikes, one before and the other one after a presynaptic spike (post-pre-post). Here, timing differences are defined as $\Delta t_1=t_{post1}-t_{pre}$ and $\Delta t_2=t_{post2}-t_{pre}$.

\subsubsection{Quadruplet protocol}\label{subsec:exp-quadruplet}
This protocol is composed of 60 quadruplets of spikes repeated at frequency of $\rho=1$~Hz. The quadruplet is composed of either a post-pre pair with a delay of $\Delta t_1=t_{post1}-t_{pre1}<0$ precedes a pre-post pair with a delay of $\Delta t_2=t_{post2}-t_{pre2}>0$ with a time $T>0$, or a pre-post pair with a delay of $\Delta t_2=t_{post2}-t_{pre2}>0$ precedes a post-pre pair with a delay of $\Delta t_1=t_{post1}-t_{pre1}<0$ with a time $T<0$, where $T=(t_{pre2}+t_{post2})/2-(t_{pre1}+t_{post1})/2$. Identical to [2], in all quadruplet experiments in this paper, $\Delta t=-\Delta t_1=\Delta t_2$.   

\subsection{Data sets}\label{subsec:data}
The simulations were conducted using two types of data sets: The first data set originates from experiments on the visual cortex~\cite{ref4} which investigated how altering the repetition frequency of spike pairings affects the overall synaptic weight change. This data set is composed of 10 data points (obtained from Table 1 of~\cite{ref2}) that represents experimental weight change,$~\Delta w$, for two different$~\Delta t$'s, and as a function of the frequency of spike pairs under a pairing protocol in the visual cortex. The second experimental data set that was utilized, originates from hippocampal cultures experiments from~\cite{ref3} which examined pairing, triplet and quadruplet protocols effects on synaptic weight change. This data set consists of 13 data points obtained from Table 2 of~\cite{ref2}. This data set shows the experimental weight change,$~\Delta w$, as a function of the relative spike timing $~\Delta t$, $~\Delta t_1$, $~\Delta t_2$ and $T$ under pairing, triplet and quadruplet protocols in hippocampal cultures. 

\subsection{Error function}\label{subsec:err}

Identical to~\cite{ref2} that tests its proposed triplet model simulation results against the experimental data and reports their differences as Normalized Mean Square Error (NMSE) for each data set, we verified our circuit simulation results under same condition. The mentioned NMSE~\cite{ref2} is calculated using the following equation:

\begin{eqnarray}\label{eq:err}
E = \frac{1}{p}\sum_{i=1}^p\left( \frac{\Delta w^i_{\rm exp}-\Delta w^i_{\rm cir}}{\sigma_i}\right)^{2}\hspace{-2mm},\label{eq:3}
\end{eqnarray}
where $\Delta w^i_{\rm exp}$, $\Delta w^i_{\rm cir}$ and $\sigma_i$ are the mean weight change obtained from biological experiments, the weight change obtained from the circuit under consideration, and the standard error mean of $\Delta w^i_{\rm exp}$ for a given data point $i$, respectively; $p$ represents the number of data points in a specified data set (can be 10 or 13).      

In order to minimize the resulting NMSE for a circuit, there was a need to adjust the parameters and time constants to minimize the resulting NMSE. In the following subsections, the circuit simulation results and applied bias currents for setting the required parameters, in order to have the minimum achieved NMSEs are reported. Both mentioned circuit shown in figures~\ref{fig:1}(b) and~\ref{fig:triplet} were simulated using parameters for a 0.35$~\mu$m SiGe CMOS process. This process was selected because of cost-related issues, rather than the high speed of this process. All transistors are 0.7$~\mu$m wide and 0.35$~\mu$m long. The capacitor values are: $C_W$=10 pF, and other capacitors are equal to 10 fF in both presented circuits. 

The synaptic weight capacitor in the proposed design occupies a large portion of the silicon area, as it is the case for almost all synaptic circuits. Therefore, one of our concerns is to reduce this capacitor value in order to make the area of the design, smaller. Some of our simulation results show that, it is possible to scale the capacitor value (and hence, its size) to one/tenth, however, the circuits bias currents need to be retuned.   

It should be noted that, during all experiments in this paper, the nearest spike interaction, which considers the interaction of a spike only with its two immediate succeeding and immediate preceding nearest neighbours, was used~\cite{ref2}. 

\section{Simulation Results}\label{sec:sim} 
\subsection{P-STDP circuit simulation results}\label{subsec:expp}
\begin{figure}
\centering
  \includegraphics[width=.5\textwidth]{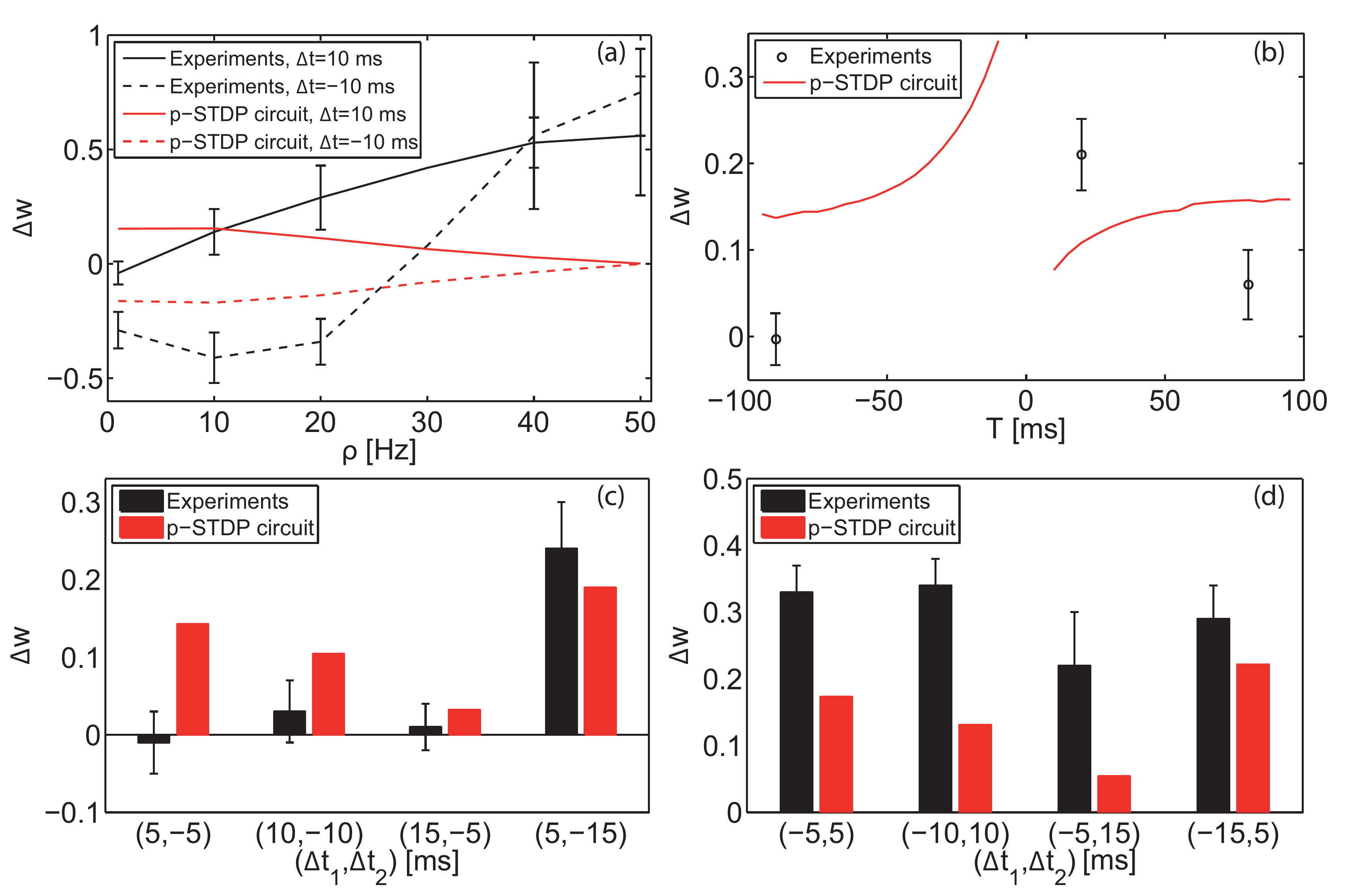}
  \caption{p-STDP circuit simulation results. Please note, there are no data at 30 Hz. The values of~ $\rho$, $T$ and $\Delta t$ are used as described in~\cite{ref2}.}\label{fig:3}
\end{figure}

Simulation results for the first data set, which reflects the weight change under a pairing protocol and as a function of the pairing frequency,$~\rho$, using the presented circuit in Fig.~\ref{fig:1}(b) are demonstrated in Fig.~\ref{fig:3}(a). This figure shows how the p-STDP circuit which acts based on the standard STDP rule, fails to reproduce the observed experimental results in visual cortex reported in~\cite{ref4}. The minimal NMSE obtained in this situation was E=7.26 which is consistent with the reported minimal achieved error using computer simulation of the p-STDP rule in Fig. 6A of~\cite{ref2}. The four required bias currents for controlling the model parameters are reported in Table~\ref{tab:1}.

Simulation results of the second data set also suggest that p-STDP rule fails to reproduce experimental results observed in hippocampal cultures. Fig.~\ref{fig:3}(b) shows the circuit simulation results along with experimental results for the quadruplet protocol, while Fig.~\ref{fig:3}(c) and Fig.~\ref{fig:3}(d) represent the triplet protocols for pre-post-pre and post-pre-post combinations of spike triplets, respectively. The minimal NMSE obtained in this situation was E=10.76, again consistent with the reported results in Fig. 6B of~\cite{ref2}. The four required bias currents for controlling the model parameters are reported in Table~\ref{tab:1}.

\begin{table} 
\centering
\caption{p-STDP circuit bias currents}\label{tab:1}
\scriptsize \begin{tabular}{c c c c c} 
\hline
Data set             & $I_{\rm pot}$ & $I_{\rm dep}$ & $I_{\rm tp}$ & $I_{\rm td}$ \\
\hline
First        & 150nA & 150nA & 24pA & 18pA \\
Second       & 410nA & 190nA & 20pA & 5pA \\
\hline
\end{tabular}
\end{table}


\subsection{T-STDP circuit simulation results}\label{subsec:expt}

In order to test the proposed t-STDP circuit under the mentioned protocols and using the two data sets, firstly full t-STDP circuit was employed. This circuit is shown in Fig.~\ref{fig:triplet} and consists of four distinct parts each of them related to one of the pair or triplet combination of spikes. However, as stated in~\cite{ref2}, during simulations we observed that only some of these combinations are really necessary and play significant roles in synaptic weight change under different protocols. So, we changed the full t-STDP circuit to minimal t-STDP circuit in correspondence to two minimal t-STDP rules in~\cite{ref2}. In these minimal rules, the inconsequential parts of the proposed full-triplet circuit are removed to have the minimal circuits.   
As it can be extracted from the last line of Table 3 in~\cite{ref2}, the minimal t-STDP rule which is capable of reproducing the expected visual cortex weight change experiments (the first data set), set pre-post and pre-post-pre spike combination amplitude parameters to zero. And it means that this rule neither require the pre-post interactions of spikes, nor the pre-post-pre interactions to take part in synaptic weight modification. So we don't need these parts also in the corresponding minimal t-STDP circuit. This circuit composed of 19 transistors (exclude the parts in the dashed and dotted boxes in the circuit presented in Fig.~\ref{fig:triplet}) and it can reproduce very similar results to the full t-STDP circuit which contains 34 transistors (Fig.~\ref{fig:triplet}). The minimum NMSE obtained for the first data set and using this first minimal t-STDP circuit was E=0.64 which is near the minimum NMSE obtained by means of computer simulations of minimal t-STDP model, E=0.34 (error value obtained from Table 3 of~\cite{ref2}). The five required bias currents for controlling the model parameters are reported in Table~\ref{tab:2}.

\begin{table}
\centering 
\caption{First t-STDP circuit bias currents}\label{tab:2}
\scriptsize \begin{tabular}{c c c c c c c} 
\hline
Data set   & $I_{\rm dep1}$ & $I_{\rm tp1}$ & $I_{\rm td1}$ & $I_{\rm pot2}$ & $I_{\rm tp2}$ \\
\hline
First      & 300nA & 40nA & 40pA  & 1.5uA  & 50pA \\
\hline
\end{tabular}
\end{table}
  
\begin{figure}
\centering
  \includegraphics[width=.5\textwidth]{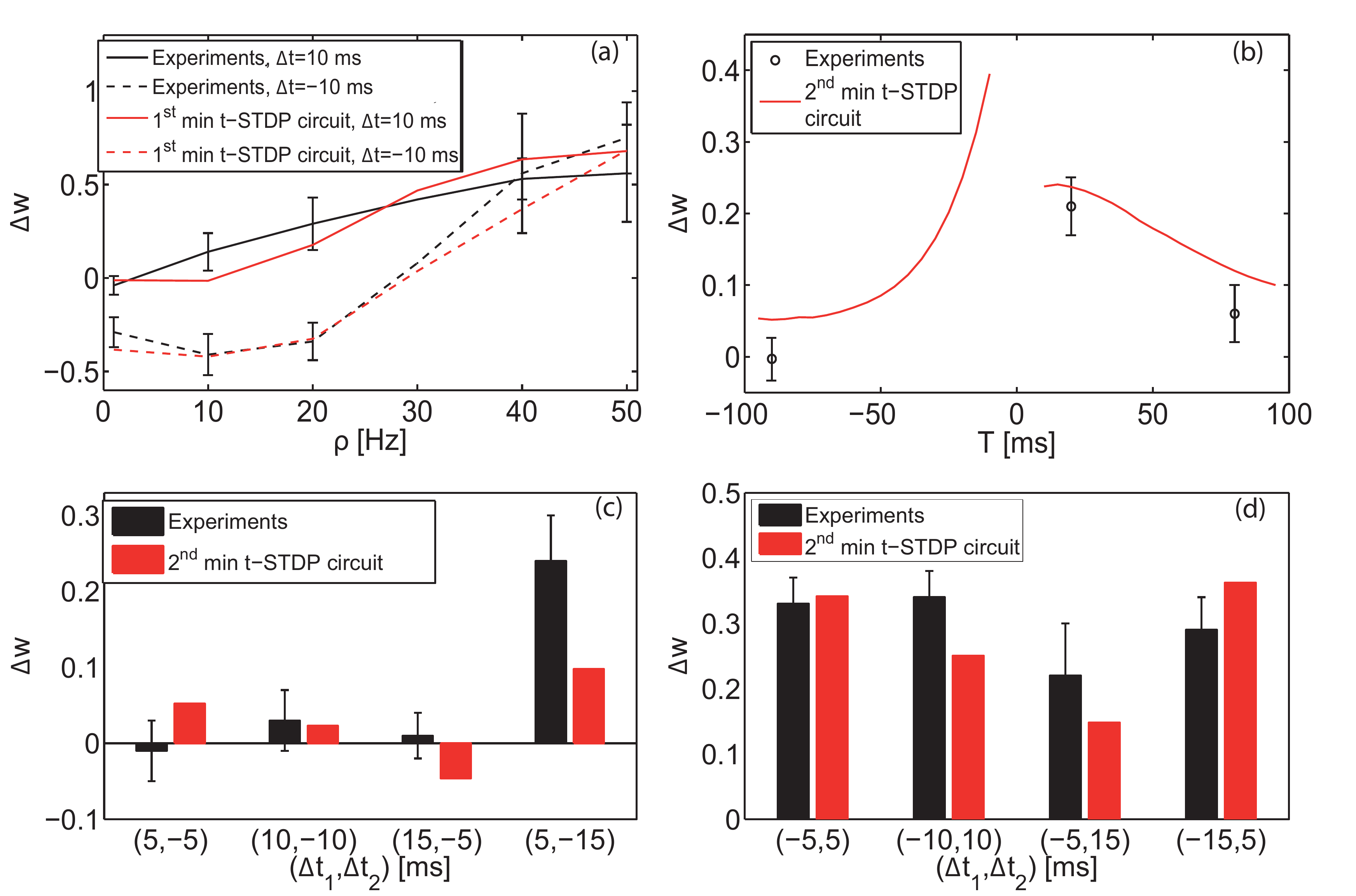}
  \caption{t-STDP circuit simulation results. Please note, there are no data at 30 Hz. The values of~ $\rho$, $T$ and $\Delta t$ are used as described in~\cite{ref2}.}\label{fig:4}
\end{figure}

Furthermore, the minimum obtained error for the second data set using the second minimal t-STDP circuit is E=2.25. The achieved results are shown in Fig.~\ref{fig:4}(b)-(d). The second minimal t-STDP circuit is composed of the whole top parts and the right bottom part of the full t-STDP circuit presented in Fig.~\ref{fig:triplet} (see the last line of Table 4 in~\cite{ref2}). The obtained NMSE using this circuit is slightly better than the NMSE obtained using minimal t-STDP model and by means of computer simulations (E=2.9 extracted from Table 4 of~\cite{ref2}). The six required bias currents for controlling the model parameters are reported in Table~\ref{tab:3}.

\begin{table}
\centering 
\caption{Second t-STDP circuit bias currents}\label{tab:3}
\scriptsize \begin{tabular}{c c c c c c c} 
\hline
Data set    & $I_{\rm pot1}$ & $I_{\rm dep1}$ & $I_{\rm tp1}$ & $I_{\rm td1}$ & $I_{\rm pot2}$ & $I_{\rm tp2}$ \\
\hline
Second      & 160nA & 130nA & 28pA  & 20pA  & 400nA & 10pA \\
\hline
\end{tabular}
\end{table}

\section{Previous works}\label{sec:prev}
There is a previous VLSI implementation~\cite{ref9}, capable of reproducing the mentioned biological experiments (except the quadruplet protocol which has not been shown in~\cite{ref9}) similar to the proposed triplet circuit. In terms of functionality, these implementations are different, since~\cite{ref9} directly implements the BCM rule, requiring the voltage of the neuron to take part in learning, and eventually leads to compatible changes in the neuron architecture. By contrast, our proposed design is based upon triplets of spikes from which we extract the required triplet, quadruplet, and pairing frequency experiments. Note that our proposed circuit is capable also to reproduce the effects of the BCM rule (results not shown but to be presented elsewhere). In addition, unlike the circuit presented in~\cite{ref9}, our VLSI circuit acts as a STDP circuit, which can be simply used to connect to other sets of (neuromorphic) neurons of choice. Furthermore, we show that the proposed circuit not only can reproduce the required behaviour seen in the mentioned experiments, but also it can be tuned to mimic those experimentally observed behaviour with a small error, while~\cite{ref9} just depicts the behaviour and not the required values observed in biological experiments. Also, the proposed circuit would require smaller silicon real states and lower power consumption when compared to the circuit presented in~\cite{ref9}.

Besides, our previous work~\cite{rahimitriplet}, presented a circuit design of t-STDP which reproduced the essential behaviour seen in the mentioned experiments except for a non-exponential learning window. This exponential learning window is an essential feature of the original pair-based and triplet-based STDP models presented in~\cite{ref11,ref2}. So in the present circuit design, the circuits closely mimic the mathematical models of STDP rules. This has lead to better synaptic weight modification capability in comparison to our initial reported design. 

\section{Discussion}\label{sec:discussion}
The proposed design uses a number of transistors operating in the subthreshold regime of operation, so it is expected that this circuit will be sensitive to process variations. However, the simulation results shown in Fig.~\ref{fig:1}(c) (a 3$\sigma$ process corner variations) depict that the time constants in the STDP learning window are almost invariant and there is a slight variation in the amplitude of the window when considering these corner variations. This can be explained by considering the fact that in these simulations similar transistors parameters are assumed for adjacent transistors.

\begin{figure}
\centering
  \includegraphics[width=.5\textwidth]{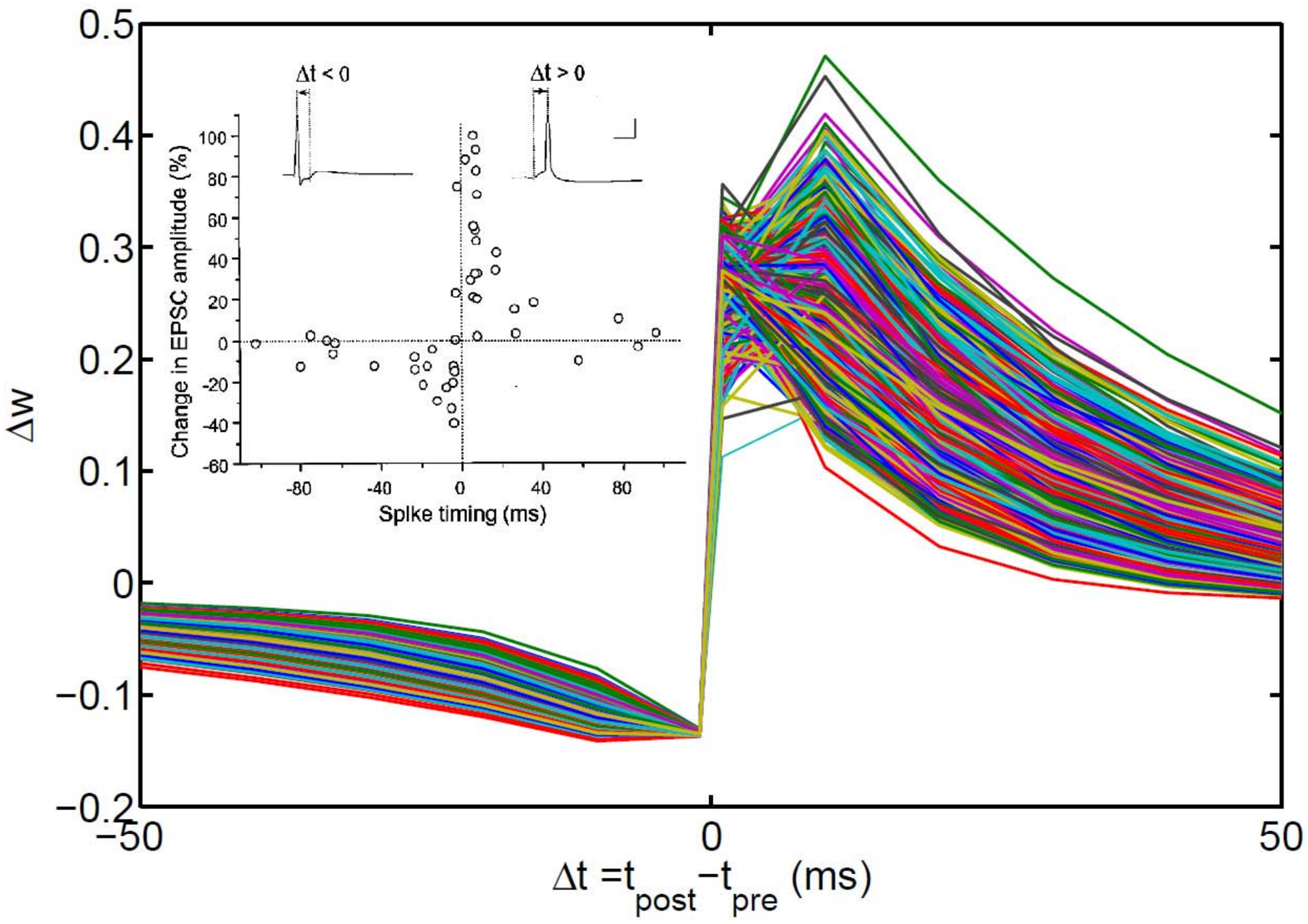}
  \caption{P-STDP circuit simulation results for 1000 Monte Carlo runs for mismatch analysis. Each curve represents the weight change for a random mismatch between current mirror transistors. The inset figure shown above is measured biological data and demonstrates the excitatory postsynaptic current (EPSC) and the noisy nature of these data~\cite{ref1}.}\label{fig:miswin}
\end{figure}

\begin{figure*} [ht!]
\centering
  \includegraphics[width=0.7\textwidth,height=1\textwidth,angle=90]{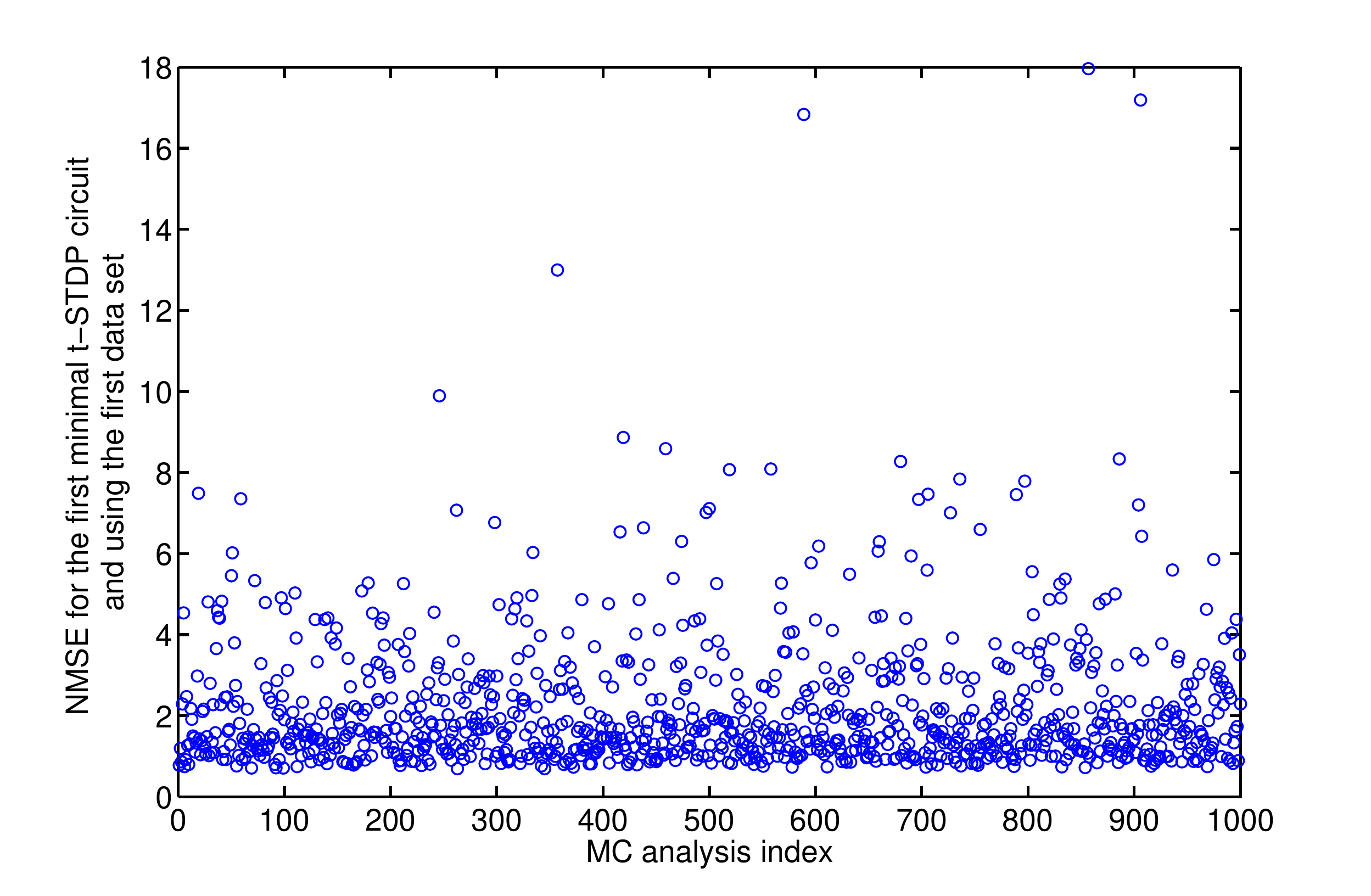}
  \caption{T-STDP circuit simulation results for 1000 Monte Carlo runs for mismatch analysis. Each run presents a NMSE value obtained from the t-STDP circuit when employing the first data set for a random mismatch between current mirror transistors.}\label{fig:NMSE}
\end{figure*}

Besides, as the circuits consist of current mirrors that are used to replicate the required amplitudes or time constants, the effect of transistors mismatch on the circuit performance must be considered. 
Here we report some preliminary results concerning the transistors mismatch in the proposed circuits using two runs of Monte Carlo (MC) analyses. These analyses were conducted on both pair- and triplet-based circuits. In these analyses, the threshold voltage was varied by using the {\tt delvto} parameter in Hspice. This parameter was set to an absolute Gaussian distribution with a nominal value of zero, absolute variation of 26.6~mV for a standard deviation of 3. These values were obtained from the MC model parameters for the mentioned process. 

The first analysis examines the effect of the current mirror transistors mismatch on the p-STDP circuit (Fig.~\ref{fig:1}(b)) when generating the pair-based STDP learning window (Fig.~\ref{fig:miswin}). This simulation shows that the design is prone to transistor mismatch and cannot reproduce the exact behaviour of the p-STDP theoretical model. However, at the same time it demonstrates that the total behaviour of Long Term Potentiation (LTP) and Long Term Depression (LTD) in the window can be regenerated. 

The second analysis was done on the t-STDP circuit, using the first data set, in the presence of current mirror transistors mismatch. Figure~\ref{fig:NMSE} shows how the NMSE for this circuit varies for 1000 various mismatch conditions created in a MC mismatch analysis. Results for the MC simulation in this case shows that the NMSE in the presence of transistor mismatch can go almost up to 18.
These simulations show a significant variation in the NMSE when compared with TT model parameter simulations.
 
Although the circuit shows a significant variation in the NMSE comparing to the NMSE in the absence of variation and using the typical model, by employing some digital trimming techniques~\cite{al2003new}, we can ensure the circuit operation with an acceptable value of NMSE. Furthermore, another method for curing this variation in the NMSE is retuning the bias currents in a way that reduce the NMSE to a minimal point even in the presence of device variations. 
The inclusion of these techniques with the proposed t-STDP circuit will be covered in a future publication.

\section{Conclusion}\label{sec:conc}
This paper presents a new VLSI circuit design for STDP learning rule based on triplets of spikes. Simulation results demonstrate that the pair-based STDP circuit cannot account for complex biological experiments i.e., triplet and quadruplet experiments, as well as the effect of pairing frequency increase on weight change, while the proposed triplet-based STDP circuit can closely mimic the outcomes observed in biological experiments. Because of these features, the proposed VLSI circuit can play a significant role in future VLSI implementation of neuromorphic systems.
  
\section*{Acknowledgment}
The support of the Australian Research Council (ARC) is gratefully acknowledged.

\bibliographystyle{ieeetran}
\bibliography{refs}

\begin{thebibliography}{10}
\providecommand{\url}[1]{#1}
\csname url@samestyle\endcsname
\providecommand{\newblock}{\relax}
\providecommand{\bibinfo}[2]{#2}
\providecommand{\BIBentrySTDinterwordspacing}{\spaceskip=0pt\relax}
\providecommand{\BIBentryALTinterwordstretchfactor}{4}
\providecommand{\BIBentryALTinterwordspacing}{\spaceskip=\fontdimen2\font plus
\BIBentryALTinterwordstretchfactor\fontdimen3\font minus
  \fontdimen4\font\relax}
\providecommand{\BIBforeignlanguage}[2]{{%
\expandafter\ifx\csname l@#1\endcsname\relax
\typeout{** WARNING: IEEEtran.bst: No hyphenation pattern has been}%
\typeout{** loaded for the language `#1'. Using the pattern for}%
\typeout{** the default language instead.}%
\else
\language=\csname l@#1\endcsname
\fi
#2}}
\providecommand{\BIBdecl}{\relax}
\BIBdecl

\bibitem{ref2}
J.~Pfister and W.~Gerstner, ``Triplets of spikes in a model of spike
  timing-dependent plasticity,'' \emph{The Journal of Neuroscience}, vol.~26,
  no.~38, pp. 9673--9682, 2006.

\bibitem{ref3}
H.~Wang, R.~Gerkin, D.~Nauen, and G.~Bi, ``Coactivation and timing-dependent
  integration of synaptic potentiation and depression,'' \emph{Nature
  Neuroscience}, vol.~8, no.~2, pp. 187--193, 2005.

\bibitem{ref5}
R.~Froemke and Y.~Dan, ``Spike-timing-dependent synaptic modification induced
  by natural spike trains,'' \emph{Nature}, vol. 416, no. 6879, pp. 433--438,
  2002.

\bibitem{ref4}
P.~Sj{\"{o}}str{\"o}m, G.~Turrigiano, and S.~Nelson, ``Rate, timing, and
  cooperativity jointly determine cortical synaptic plasticity,''
  \emph{Neuron}, vol.~32, no.~6, pp. 1149--1164, 2001.

\bibitem{ref6}
A.~Bofill-I-Petit and A.~Murray, ``Synchrony detection and amplification by
  silicon neurons with {STDP} synapses,'' \emph{IEEE transactions on neural
  networks/a publication of the IEEE Neural Networks Council}, vol.~15, no.~5,
  pp. 1296--1304, 2004.

\bibitem{ref1}
G.~Bi and M.~Poo, ``Synaptic modifications in cultured hippocampal neurons:
  dependence on spike timing, synaptic strength, and postsynaptic cell type,''
  \emph{The Journal of Neuroscience}, vol.~18, no.~24, pp. 10\,464--10\,472,
  1998.

\bibitem{ref7}
G.~Indiveri, E.~Chicca, and R.~Douglas, ``A {VLSI} array of low-power spiking
  neurons and bistable synapses with spike-timing dependent plasticity,''
  \emph{IEEE Transactions on Neural Networks}, vol.~17, no.~1, pp. 211--221,
  2006.

\bibitem{ref8}
K.~Cameron, V.~Boonsobhak, A.~Murray, and D.~Renshaw, ``Spike timing dependent
  plasticity ({STDP}) can ameliorate process variations in neuromorphic
  {VLSI},'' \emph{IEEE Transactions on Neural Networks}, vol.~16, no.~6, pp.
  1626--1637, 2005.

\bibitem{ref9}
C.~Mayr, M.~Noack, J.~Partzsch, and R.~Schuffny, ``Replicating experimental
  spike and rate based neural learning in {CMOS},'' in \emph{Proceedings of
  IEEE International Symposium on Circuits and Systems (ISCAS)}, 2010, pp.
  105--108.

\bibitem{rahimitriplet}
M.~Rahimi~Azghadi, O.~Kavehei, S.~Al-Sarawi, N.~Iannella, and D.~Abbott,
  ``Novel {VLSI} implementation for triplet-based spike-timing dependent
  plasticity,'' in \emph{Proceedings of the 7th International Conference on
  Intelligent Sensors, Sensor Networks and Information Processing}, 2011, pp.
  158--162.

\bibitem{ramakrishnan2011floating}
S.~Ramakrishnan, P.~Hasler, and C.~Gordon, ``Floating gate synapses with
  spike-time-dependent plasticity,'' \emph{IEEE Transactions on Biomedical
  Circuits and Systems}, vol.~5, no.~3, pp. 244--252, 2011.

\bibitem{ref10}
H.~Tanaka, T.~Morie, and K.~Aihara, ``A {CMOS} spiking neural network circuit
  with symmetric/asymmetric {STDP} function,'' \emph{IEICE Transactions on
  Fundamentals of Electronics, Communications and Computer Sciences}, vol.
  {E92-A}, no.~7, pp. 1690--1698, 2009.

\bibitem{ref11}
S.~Song, {K.D. Miller}, and {L.F. Abbott}, ``Competitive {Heb an} learning
  through spike-timing-dependent synaptic plasticity,'' \emph{Nature
  Neuroscience}, vol.~3, pp. 919--926, 2000.

\bibitem{al2003new}
S.~Al-Sarawi, ``New efficient offset voltage cancellation techniques using
  digital trimming,'' in \emph{Proceedings of SPIE}, vol. 5117, 2003, pp.
  65--76.

\end{thebibliography}

\end{document}